\pdfoutput=1

\documentclass[11pt]{article}

\usepackage[]{acl}

\usepackage{times}
\usepackage{latexsym}

\usepackage[T1]{fontenc}

\usepackage[utf8]{inputenc}

\usepackage{microtype}

\usepackage{url}
\usepackage{multirow}
\usepackage{adjustbox}
\usepackage{latexsym}
\usepackage{todonotes}
\usepackage{array}
\usepackage{subfigure} 
\usepackage{algorithm}
\usepackage{algorithmic}
\usepackage{amsmath} 
\usepackage{amsfonts,amssymb}
\usepackage[normalem]{ulem}
\useunder{\uline}{\ul}{}

\title{Learning to Execute Actions or Ask Clarification Questions}

\author{Zhengxiang Shi  \and Yue Feng  \and Aldo Lipani  \\
  University College London, London, UK \\
  \texttt{\{zhengxiang.shi.19,yue.feng.20,aldo.lipani\}@ucl.ac.uk} \\}

\begin{document}
\maketitle
\begin{abstract}
Collaborative tasks are ubiquitous activities where a form of communication is required in order to reach a joint goal. 
Collaborative building is one of such tasks. 
We wish to develop an intelligent builder agent in a simulated building environment (Minecraft) that can build whatever users wish to build by just talking to the agent. In order to achieve this goal, such agents need to be able to take the initiative by asking clarification questions when further information is needed.
Existing works on Minecraft Corpus Dataset only learn to execute instructions 
neglecting the importance of asking for clarifications.
In this paper, 
we extend the Minecraft Corpus Dataset by annotating all builder utterances into eight types, including clarification questions, and propose a new builder agent model capable of determining when to ask or execute instructions.
Experimental results show that our model achieves state-of-the-art performance on the collaborative building task with a substantial improvement. We also define two new tasks, the learning to ask task and the joint learning task.
The latter consists of solving both collaborating building and learning to ask tasks jointly.

\end{abstract}

\section{Introduction}
Following instructions in natural language by intelligent agents to achieve a shared goal with the instructors
in a pre-defined environment is a ubiquitous task in many scenarios, e.g., finding a target object in an environment~\cite{nguyen2019help,roman2020rmm}, drawing a picture~\cite{lachmy2021draw}, or building a target structure~\cite{narayan2019collaborative}. 
A number of machine learning (ML) research projects about following instructions tasks have been initiated by making use of the video game Minecraft~\cite{johnson2016malmo, shu2017hierarchical, narayan2019collaborative, guss2019minerl, jayannavar2020learning}. 
Building such agents requires to make progress in
\textit{grounded natural language understanding}
-- understanding complex instructions, for example, with spatial relations in natural language -- \textit{self-improvement} 
-- studying how to flexibly learn from human interactions -- 
\textit{synergies of ML components} 
-- exploring the integration of several ML and non-ML components to make them work together~\cite{szlam2019build}.

\definecolor{wong_1}{rgb}{0.9019, 0.6235, 0}
\definecolor{wong_4}{rgb}{0,0.6196,0.4509}

\begin{figure}[!t]
  \centering
  \includegraphics[width=0.5\textwidth]{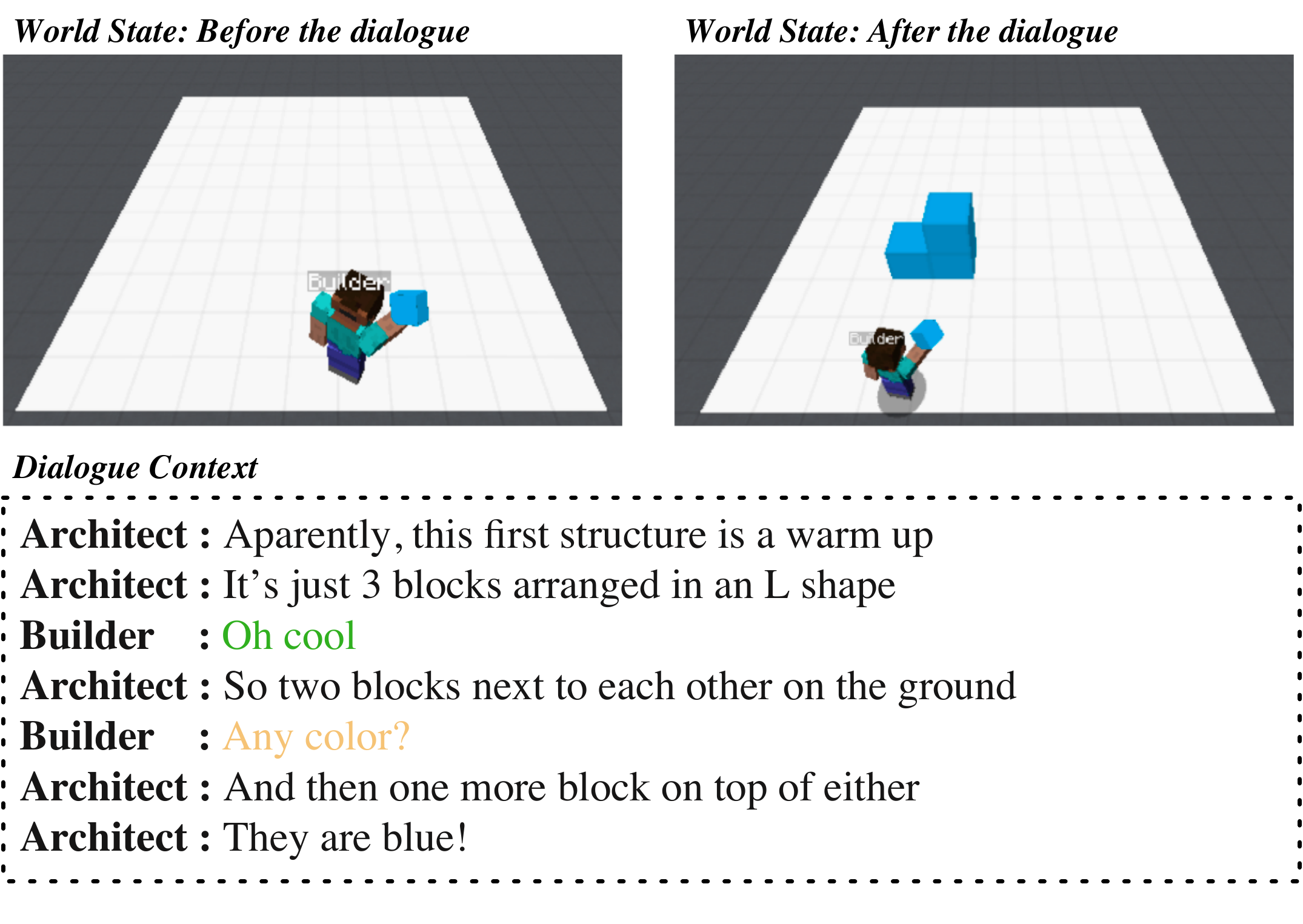}
  \caption{A simple example of builder task: The builder can observe the world state and dialogue context. For the sake of space, only a part of the dialogue history is displayed. 
  The utterance in {\color{wong_4}green} displays understanding and 
  the utterance in {\color{wong_1}yellow} asks a clarification question. 
  }
  \label{fig:task}
\end{figure}

The recently introduced Minecraft Corpus dataset~\cite{narayan2019collaborative} proposes a collaborative building task, in which an architect and a builder can communicate via a textual chat. Architects are provided with a target structure they want to have built, and the builders are the only ones who can control the Minecraft avatar in the virtual environment. The task consists in building 3D structures in a block world-like scenario collaboratively, as shown in the Figure \ref{fig:task}.
Earlier works in Minecraft collaborative building tasks~\cite{jayannavar2020learning} attempted to build an automated builder agent with a large action space but failed to allow the builder to take the initiative in the conversation.
However, an intelligent agent should not only understand and execute the instructor's requests but also be able to take initiatives, e.g., asking clarification questions, in case the instructions are ambiguous.
In the task defined by this dataset, builders may encounter ambiguous situations that are hard to interpret by just relying on the world state information and instructions. For example, in Figure \ref{fig:task}, we provide a simple case where the architect fails to provide sufficient information to the builder, such as the color of the blocks. In this situation, it is clearly difficult for the builder to know exactly which action should be taken. If, however, the builder is able to clarify the situation with the architect, this ambiguity can be resolved. Therefore, builders, besides following architects' instructions, should take the initiative in the conversation and ask questions when necessary. %

To this end, in this paper we annotate all builder utterances in the Minecraft Corpus dataset by categorizing them in the dataset into eight dialogue utterance types as shown in Table \ref{tbl:statistics}, allowing the intelligent agents to learn when and what to ask given the world state and dialogue context. Particularly, a builder would ask \textit{task-level questions} or \textit{instruction-level questions} for further clarifications. 
Experimental results in the Sec.~\ref{sec:learn_to_ask} show that determining when to ask clarification questions remains a challenging task. However, it is worth noting that the clarification questions in the Minecraft Corpus dataset are more complex and diverse than those in navigation tasks~\cite{roman2020rmm,nguyen2019help,thomason2020vision} whose questions are relatively simpler and mainly about where to go.

Also, we propose a new automated builder agent that learns to map instructions to actions and decide when to ask questions. Our model utilizes three dialogue slots, the action type slot, the location slot, and the color slot. This solution has the benefit of making the learning easier with respect to those models that work using a large action space ~\cite{jayannavar2020learning}. 
To solve the collaborative building task, both the dialogue context and the world state need to be considered. Therefore, to endow our model with the ability to better learn the representations between the world state and language, our model implements a cross-modality module, which is based on the cross attention mechanism.
Experimental results on our extended Minecraft Corpus dataset show that our model achieves state-of-the-art performance with a substantial improvement for the \textit{collaborative building task}. We also provide new baselines for \textit{learning to ask} task and the joint learning of these two tasks.


\section{Related Work and Background}
\paragraph{Dialogue Tasks.}
As virtual personal assistants have now penetrated the consumer market, with products such as Siri and Alexa, the research community has produced several works on \textit{task-oriented dialogue tasks} such as: 
hotel booking, 
restaurant booking, 
movie recommendation, etc.~\cite{budzianowski2018multiwoz, wei2018airdialogue, wu2019transferable,feng2021sequence,feng2022dynamic,kim-user-simulation}.
These task-oriented dialogues have been modelled as slot filling tasks. These tasks consist of correctly identifying and extracting information (slots) useful to solve the task. 
However, most of these slot filling tasks~\cite{coope2020span,ni2020natural} are considered as semantic tagging or parsing of natural language and do not normally consider visual information. Moreover, these tasks focus only on two of the many components needed by conversational systems: the Natural Language Understanding (NLU) and Dialogue State Tracking (DST) ones~\cite{budzianowski2018multiwoz,williams2014dialog}. 
Beside these task-oriented dialogue tasks, the research community has also focused on \textit{instruction following dialogue tasks}, such as
target completion tasks~\cite{de2017guesswhat}, 
object finding tasks~\cite{roman2020rmm}, and navigation tasks~\cite{thomason2020vision}. 
\citet{narayan2019collaborative} proposed the Minecraft Corpus dataset, where the task consists in a cooperative asymmetric task involving an architect and a builder that have to build a target structure collaboratively. %
\citet{jayannavar2020learning} then built a builder model to follow the sequential instructions from the architect.

\paragraph{Multi-Modal.}
Almost all instruction following dialogue tasks need to consider both contextual information and actions as well as the state of the world~\cite{suhr2018situated, suhr2019executing, chen2019touchdown, lachmy2021draw}, which remains a key challenge for instruction following dialogue tasks. %
In particular, the Vision-and-Dialog Navigation (VDN) task~\cite{chen2019touchdown, thomason2020vision,roman2020rmm, zhu2021self} where the question-answering dialogue and visual contexts are leveraged to facilitate navigation, has attracted increasing research attention. %
Other tasks, such as moving blocks tasks~\cite{misra2017mapping} and object finding tasks~\cite{janner2018representation}, 
also require the modelling of both contextual information in natural language as well as the world state representation to be solved.

\begin{table*}[!ht]
\centering
\small
\caption{The taxonomy of builder utterances: we categorize them into eight types where \textit{instruction-level questions} and \textit{task-level questions} are both a sub-type of clarification questions. %
There are 4${\small,}$904 builder utterances in total.}
\begin{tabular}{llll}
\hline
\hline
\multicolumn{1}{c}{\textbf{Category}}  & \multicolumn{1}{c}{\textbf{Example}}                                                                                                                                 & \multicolumn{1}{c}{\textbf{Amount}} & \multicolumn{1}{c}{\textbf{Percentage}} \\ \hline
{\ul \textit{Instruction-level Questions}} & \begin{tabular}[c]{@{}l@{}}1. What color?\\ 2. Is it flat?\end{tabular}                                                                                              & 914                                 & 18.64\%                           \\ \hline
{\ul \textit{Task-level Questions}}         & \begin{tabular}[c]{@{}l@{}}1. What are we building?\\ 2. What's next?\end{tabular}                                                                                   & 252                                 & 5.14\%                            \\ \hline
{\ul \textit{Verification Questions}}  & \begin{tabular}[c]{@{}l@{}}1. Like that or othwr way?\\ 2. Is this the cross you wanted?\end{tabular}                                                                & 1021                                & 20.82\%                           \\ \hline
{\ul \textit{Greeting}}                & \begin{tabular}[c]{@{}l@{}}1. Ready?\\ 2. Hello!\end{tabular}                                                                                                        & 808                                 & 16.48\%                           \\ \hline
{\ul \textit{Suggestions}}             & \begin{tabular}[c]{@{}l@{}}1. In the future we can call these donuts or something\\ 2. No problem, if it's hard to describe we can just go step by step\end{tabular} & 59                                  & 1.23\%                            \\ \hline
{\ul \textit{Display Understanding}}   & \begin{tabular}[c]{@{}l@{}}1. No problem.\\ 2. Knew what you meant\end{tabular}                                                                                      & 1296                                & 26.43\%                           \\ \hline
{\ul \textit{Status Update}}           & \begin{tabular}[c]{@{}l@{}}1. I don't have enough green to continue.\\ 2. I'll stay with this perspective\end{tabular}                                               & 101                                 & 2.06\%                            \\ \hline
{\ul \textit{Others}}    & \begin{tabular}[c]{@{}l@{}}1. I got my first job from Minecraft.\\ 2. Oh, wwo, sorry!\end{tabular}                                                                   & 453                                 & 9.24\%                            \\ \hline
\hline
\end{tabular}
\label{tbl:taxonomy}
\end{table*}

\paragraph{Spatial Reasoning.} Many instruction following dialogue tasks contain texts with spatial-temporal concepts~\cite{chen2019touchdown, yang2020robust}. For instance, the Minecraft Corpus dataset contains utterances with spatial relations, e.g., ``go to the middle and place an orange block two spaces to the left''. 
Although pre-trained language models have been used successfully in a wide array of downstream tasks,
interpreting and grounding abstractions stated in natural language, such as spatial relations, have not been systematically studied and remain still challenging. %
Therefore, another challenge for an agent is to follow instructions which require the learning and understanding of spatio-temporal linguistic concepts in natural language. 
To train models able to
understand and reason about spatial references in natural language,~\citet{shi2022stepgame} proposed a benchmark for robust multi-hop spatial reasoning over texts. 

\paragraph{Learning by Asking Questions.} 
Determining whether to ask clarification questions and what to ask is critical for instruction followers to complete the tasks. Several recent studies have focused on learning a dialogue agent with the ability to interact with users by both responding to questions and by asking questions to accomplish their task interactively~\cite{li2016learning,de2017guesswhat,misra2018learning,roman2020rmm}. For instance, \citet{de2017guesswhat} introduced a game to locate an unknown object via asking questions about objects in a given image.
A decision-maker is introduced to learn when to ask questions by implicitly reasoning about the uncertainty of the agent. Different from earlier works~\cite{kitaev2017misty,suhr2019executing}, recent works on VDN tasks propose agents that learn to ask a question when the certainty of the next action is low~\cite{thomason2020vision,roman2020rmm,chi2020just}.
\citet{roman2020rmm} proposed a two models-based agent with a navigator model and a questioner model. The former model was responsible for moving towards the goal object, while the latter model was used to ask questions. 
\citet{zhu2021self} proposed an agent that learned to adaptively decide whether and what to communicate with users in order to acquire instructive information to help the navigation.

\section{Dataset and Tasks}
\subsection{The Minecraft Dialogue Corpus}
The Minecraft Dialogue Corpus~\cite{narayan2019collaborative} is built upon a simulated block-world environment with dialogues between an architect and a builder. This consists of 509 human-human dialogues (15${\small,}$926 utterances, 113${\small,}$116 tokens) playing the role of an architect and a builder, and game logs for 150 target structures of varying complexity (min.~6 blocks, max.~68 blocks, avg.~23.5 blocks), 
For each target structure at least three dialogues are collected where each dialogue contains 30.7 utterances (22.5 architect utterances and 8.2 builder utterances) and 49.5 builder blocks movements on average.

The architect instructs about a target structure the builder to build it via a dialogue. Although the architect observes the builder operating in the world, only the builder can move blocks. The builder has access to an inventory of 120 blocks of six given colors that he or she can place or remove. The collaborative building task restricts the structures to a build region of size 11 × 9 × 11, and contains 3709, 1331, and 1616 samples for training, validation, and test sets.

\subsection{Builder Dialogue Annotation}
\begin{table}[ht!]
\centering
\small
\caption{Statistics of the extended Minecraft Dialogue Corpus: "Execution (Original)" represents that the builder should predict a sequence of building actions given the dialogue context and the world state in a sample; "Ask for clarifications" indicates that the builder should ask for more information in order to execute building actions; "Others" stands for remaining dialogue acts for the builder such as greetings, chit-chat, and display understanding.}
\begin{adjustbox}{max width=\textwidth}
\begin{tabular}{l|rrr}
\hline
\hline
 & \multicolumn{1}{c}{Train} & \multicolumn{1}{c}{Valid} & \multicolumn{1}{c}{Test} \\ \hline
Execution (Original) & 3709 & 1331 & 1616 \\
Ask for clarifications & 437 & 151 & 163 \\
Others & 837 & 267 & 366 \\ \hline
Total & 4983 & 1749 & 2145 \\ \hline
\hline
\end{tabular}
\end{adjustbox}
\label{tbl:statistics}
\end{table}
Builders need to be able to decide their actions at any time point rather than only execute actions with the information about when to execute. 
Thus, we annotate all builders' utterances in the Minecraft Corpus dataset~\cite{narayan2019collaborative} and categorize all 4${\small,}$904 builder utterances into 8 utterance types. These types are partially inspired by the work of~\citet{lambertvirtual}. 
Each utterance falls into exactly one category. These categories are defined as follows:
\textit{\textbf{(1) Instruction-level Questions:}} used to request that the architect clarifies previous instructions;
\textit{\textbf{(2) Task-level Questions:}} used to request the architect to give a description about the whole picture of the building task, e.g., ask for the next instruction or ask to describe how the target structure should look like;
\textit{\textbf{(3) Verification Questions:}} used to request the architect to give feedback on previous actions;
\textit{\textbf{(4) Greetings:}} used to greet each other;
\textit{\textbf{(5) Suggestions:}} used to provide suggestions about building;
\textit{\textbf{(6) Display Understanding:}} used to express whether previous instructions have been understood;
\textit{\textbf{(7) Status Update:}} used to describe the current status, e.g., tell the architect where they are, their current block stock status, or whether they have finished a given instruction;
\textit{\textbf{(8) Others:}} not relevant to the task itself (e.g. chit-chat, expressing gratification, or apologies).

Among these 8 utterance types, the \textit{instruction-level questions} and the \textit{task-level questions} are a sub-type of clarification questions used to further clarify instructions or the task itself when the information from the architect is not clear or ambiguous. 
Based on these annotations, we extend the original dataset (the first row in Table \ref{tbl:statistics}) with two other dialogue acts, 'Ask for clarifications' and 'Others', as shown in the second and third rows of Table \ref{tbl:statistics}. Each 'Ask for clarifications' sample includes a world state and a dialogue context followed by a builder utterance labelled as \textit{instruction-level questions} or \textit{task-level questions};
each sample tagged as 'Others' includes a world state and a dialogue context followed by other builder utterance types.

\begin{figure*}[!th]
  \centering
  \includegraphics[width=\textwidth]{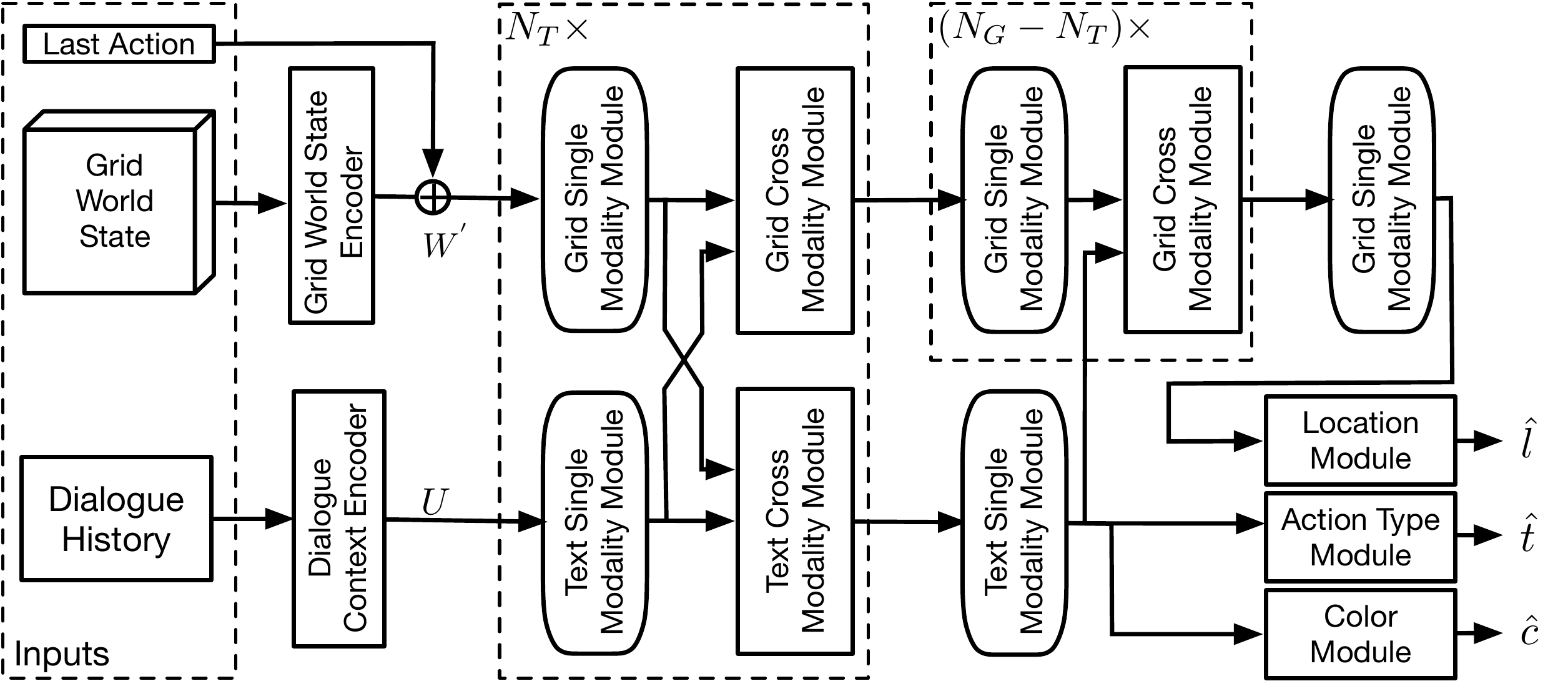}
  \caption{The model architecture. The $\oplus$ sign represents the concatenation operation. This illustration uses the plate notation. There are a total of $N_T+1$ text single modality modules, $N_G+1$ grid single modality modules, $N_T$ text cross modality modules, and $N_T$ grid cross modality modules. Arrows indicate the flow of information.}
  \label{fig:model}
\end{figure*}

\subsection{Task Definition}

Let $\mathcal{H}$ be the set of all dialogue contexts, %
$\mathcal{W}$ the set of all world states, and %
$\mathcal{A}$ the set of actions, including building actions (placing a block, removing a block, or a special stop action that terminates the task) and utterance actions (ask clarification questions or other utterance categories).
Given a dialogue context $h \in \mathcal{H}$, an initial grid-based world state $w_0 \in \mathcal{W}$, the target is to predict the next action type as 
\textit{Execution}, \textit{Ask for clarifications} and \textit{Others}.
When the prediction of the action type is \textit{Execution}, 
a sequence of actions $\{a_i\}_{i=1}^{n}$, where $a_i \in \mathcal{A}$ and $a_n$ is the \textit{Stop} action, is needed to be generated to reach the final target structure $w_n$.
The action type \textit{Execution} will update the world state via a deterministic transition function $T$ such that $w_{i} = T(w_{i-1},a_i)$ where $w_i \in \mathcal{W}$.

\section{Method}
In this section we introduce the proposed builder model, as shown in Figure~\ref{fig:model}. 
The model comprises four major components: the \textit{utterance encoder}, the \textit{world state encoder}, the \textit{fusion module}, and the \textit{slot decoder}.
The utterance encoder (in Sec.~\ref{sec:text_encoder}) and world state encoder (in Sec.~ \ref{sec:world_encoder}) learn to represent the dialogue context and the world state.  
These encoded representations are then fed into the fusion module (in Sec.~\ref{sec:fusin_module}) that learns contextualized embeddings for the grid world and textual tokens through the single and cross modality modules. Finally, the learned world and text representations are mapped into the pre-defined slot-values in the slot decoder (in Sec.~\ref{sec:decoder}).

\subsection{Dialogue Context Encoder}
\label{sec:text_encoder}
We add ``architect'' and ``builder'' annotations before each architect utterance $A_t$ and each builder utterance $B_t$ respectively. Then, the dialogue utterances are represented as 
$$D_t = \text{``architect''} A_t \oplus \text{``builder''} B_t$$ at the turn $t$, where $\oplus$ is the operation of sequence concatenation.
The entire dialogue context is defined as:
\begin{align}
    \footnotesize
    H = D_1 \oplus D_2 \oplus \dots \oplus D_t
\end{align}
Given the dialogue context $H$, we truncate the tokens from the end of the dialogue context or pad them to a fixed length as inputs and then use the dialogue context encoder to encode utterance history into 
$U \in \mathbb{R}^{s \times d_w}$, where $d_w$ is the dimension of the word embedding and $s$ is the maximum number of tokens for a dialogue context. The dialogue context encoder can be word embeddings like Glove~\citet{pennington2014glove} or contextual word embeddings~\citet{devlin2018bert}, which are both widely used in the literature~\cite{ni2021staresgru,ni2021hybrid,10.1145/3340531.3411978,10.1145/3442381.3450038}.

\subsection{Grid World State Encoder}
\label{sec:world_encoder}

The world state is represented by a voxel-based grid. 
We first represent each grid state as a 7-dimensional one-hot vector that stands for empty or a block having one of six colors, yielding a 7×11×9×11 world state representation. Additionally, we truncate the action history to the last five ones, assign an integer weight in $1,\dots,5$ and then include these weights as a separate input feature in each grid, resulting in a raw world state input of $W_{0} \in \mathbb{R}^{8 \times 11 \times 9 \times 11}$. 
We also represent the last action as an 11-dimensional vector $a$ where the first two dimensions represent the placement or removal actions, the next six dimensions represent the color, and the last three dimensions represent the location of the last action.

The structure of the world state encoder is similar to \citet{jayannavar2020learning}'s, i.g., consisting of $k$ 3D-convolutional layers ($f_{1}$) with kernel size 3, stride 1 and padding 1, followed by a ReLU activation function. 
Between every successive pair of these layers there is a 1×1×1 3D-convolutional layer ($f_{2}$) with stride 1 and no padding followed by ReLU:
\begin{align}%
    & W_{i} = \text{ReLU}(f^{i}_{2}(\text{ReLU}(f^{i}_{1}(W_{i-1})))), \\
    & W_{k} = \text{ReLU}(f^{i}_{1}(W_{k-1})),       
\end{align}
where $i = 1, 2, \dots , k-1$. $W_{k} \in \mathbb{R}^{d_c \times 11 \times 9 \times 11}$ is the learned world grid-based representation where $d_c$ is the dimension of each grid representation. Then we concatenate the last action representation $a \in \mathbb{R}^{11}$ to each grid vectors in $W_{k}$ and reshape them into $W^{'} \in \mathbb{R}^{d_c^{'} \times 1089}$, where $d_c^{'} = d_c+11$.

\subsection{Fusion Module}
\label{sec:fusin_module}

The fusion module comprises four major components: two \textit{single modality modules} and two \textit{cross-modality modules}. The former modules are based on self-attention layers and the latter on cross-attention layers. These take as input the world state representation and dialogue history representation. 
Between every successive pair of grid single-modality modules or text single-modality modules there is a cross modality module. We take $N_G$ and $N_T$ layers for the grid cross modality module and the text cross modality module.
We first revisit the definition and notations about the attention mechanism~\cite{bahdanau2014neural} and then introduce how they are integrated into our single modality modules and cross-modality modules.
\paragraph{Attention Mechanism.}
Given a query vector $x$ and a sequence of context vectors $\{y_j\}^{K}_{j=1}$, the attention mechanism first computes the matching score $s_j$ between the query vector $x$ and each context vector $y_j$. Then, the attention weights are calculated by normalizing the matching score: %
$a_j = \frac{exp(s_j)}{\sum_{j=1}^{K}exp(s_j)}$. %
The output of an attention layer is the attention weighted sum of the context vectors: %
$Attention(x, {y_j}) = \sum_{j} a_j \cdot y_j$. %
Particularly, the attention mechanism is called self-attention when the query vector itself is in the context vectors $\{y_j\}$. We use the multi-head attention following~\citet{devlin2018bert, tan2019lxmert}.
\paragraph{Single-Modality Module.} 
Each layer in a single-modality module contains a self-attention sub-layer and a feed-forward sub-layer, where the feed-forward sub-layer is further composed of a linear transformation layer, a dropout layer and a normalization layer. 
We take $N_G+1$ and $N_T+1$ layers for the grid single-modality modules and the text single-modality modules respectively, interspersed with cross-modality module as shown in Figure \ref{fig:model}. Since new blocks can only be feasibly placed if one of their faces touches the ground or another block in the Minecraft world, we add masks to all infeasible grids in the grid single-modality modules. For a set of text vectors $\{u_i^{n}\}_{i=1}^{s}$ and a set of grid vectors $\{w_j^{m}\}_{j=1}^{1089}$ as inputs of $n$-th text single-modality layer and $m$-th grid single-modality layer, where $n\in \{1,\dots,N_T+1\}$ and $m \in \{1,\dots,N_G+1\}$, we first feed them into two self attention sub-layers:
\begin{align}%
    u_i^{n} &= \text{SelfAttn}_u^{n}(u_i^{n}, \{u_i^{n}\}), \\
    w_j^{m} &= \text{SelfAttn}_w^{m}(w_j^{m}, \{w_j^{m}\}, mask)
\end{align}
Lastly, the outputs of self attention modules, $u_i^{n}$ and $w_j^{m}$, are followed by feed-forward sub-layers to obtain $\hat{u_i}^{n}$ and $\hat{w_j}^{m}$.
\paragraph{Cross-Modality Module.} 
Each layer in the cross-modality module consists of one cross-attention sub-layer and one feed-forward sub-layer, where the feed-forward sub-layers follow the same setting as the single-modality module. 
Given the outputs of $n$-th text single-modality layer, $\{\hat{u_i}^{n}\}_{i=1}^{s}$, and the $m$-th grid single-modality layer, $\{\hat{w_j}^{m}\}_{j=1}^{1089}$, as the query and context vectors, we pass them through cross-attention sub-layers, respectively:
\begin{align}%
    \hat{u_i}^{n+1} &= \text{CrossAttn}_u^{n}(\hat{u_i}^{n}, \{\hat{w_j}^{m}\}), \\
    \hat{w_j}^{m+1} &= \text{CrossAttn}_w^{m}(\hat{w_j}^{m}, \{\hat{u_i}^{n}\}),
\end{align}
The cross-attention sub-layer is used to exchange the information and align the entities between the two modalities in order to learn joint cross-modality representations. Then the output of the cross-attention sub-layer is processed by one feed-forward sub-layer to obtain $\{u_i^{n+1}\}_{i=1}^{s}$ and $\{w_j^{m+1}\}_{j=1}^{1089}$, which will be passed to the following singe-modality modules.

Finally, we obtain a set of word vectors, $\{\hat{u_i}^{N_T+1}\}_{i=1}^{s}$, and a set of grid vectors, $\{\hat{w_j}^{N_G+1}\}_{j=1}^{1089}$, that is, $U^{N_T}$ and $W^{N_G}$.
Since the value of $N_G$ and $N_T$ could be different, the modality with more layers would keep using the last single modality module's output of another modality as the input of its cross modality modules, as shown in the Figure \ref{fig:model}.

\subsection{Slot Decoder}
\label{sec:decoder}
The Slot Decoder contains three linear projection layers of trainable parameters, $M_{L} \in \mathbb{R}^{d_c^{'}}, M_{C} \in \mathbb{R}^{6 \times d_w}, M_{T} \in \mathbb{R}^{d_a \times d_w}$ where $d_a$ is the number of action types to predict. 
We compute the average of $U^{N_T} \in \mathbb{R}^{s \times d_w}$ alongside the $s$-dimension to obtain $u \in \mathbb{R}^{d_w}$.
Then we compute location logits, color logits, and action type logits: 
\begin{align}%
    \hat{l} &= \text{softmax}(M_{L} \cdot W^{N_G}), \\
    \hat{c} &= \text{softmax}(M_{C} \cdot u), \\
    \hat{t} &= \text{softmax}(M_{T} \cdot u),
\end{align}
where softmax functions are used to map the extracted information into $\hat{l} \in \mathbb{R}^{1089}$, $\hat{c} \in \mathbb{R}^{6}$, and $\hat{t} \in \mathbb{R}^{d_a}$. 

\section{Experiment, Results and Discussion}
\label{sec:experiment}
In this section, we first compare our model against the baseline for the collaborative building task where models only need to learn the instruction following task (in Sec.~\ref{sec:collaborative}). Then, we train our model to learn when to ask and evaluate on our extended Minecraft Dialogue Corpus (in Sec.~\ref{sec:learn_to_ask}). Finally, we evaluate our model's ability on the combination of the two above-mentioned tasks (in the Sec.~\ref{sec:joint_task}).
All training details are reported in the Appendix. The software and data are available at:{~\url{https://github.com/ZhengxiangShi/LearnToAsk}}.

\begin{table}[ht!]
\small
\centering
\caption{Evaluation on the collaborative building task. }
\begin{adjustbox}{max width=\textwidth}
\begin{tabular}{r|c|cccc}
\hline
\hline
\multirow{2}{*}{Model} & 
\multirow{2}{*}{Metric} & 
\multicolumn{4}{c}{Augmentation} \\ \cline{3-6}
 & & None & 2x & 4x & 6x \\ \hline \hline
\multirow{3}{*}{BAP model} & F1        & 16.1   & 18.7 & 18.4 & 18.2 \\
                           & Recall    & 12.6   & 15.2 & 14.3 & 15.0 \\
                           & Precision & 22.4   & 24.5 & 25.7 & 23.3 \\ \hline
\multirow{3}{*}{Ours (GloVe)} & F1 & \textbf{35.0} & \textbf{36.5} & \textbf{37.8} & \textbf{39.4} \\
 & Recall & 28.3 & 30.1 & 31.4 & 33.4 \\
 & Precision & 45.8 & 46.2 & 47.6 & 48.1 \\ \hline
\multirow{3}{*}{Ours (BERT)} & F1 & 34.5 & 30.1 & 30.4 & 35.4 \\
 & Recall & 26.7 & 23.6 & 23.4 & 27.9 \\
 & Precision & 48.7 & 42.6 & 43.5 & 48.5 \\ \hline
\hline
\end{tabular}
\end{adjustbox}
\label{tbl:oracle}
\end{table}

\subsection{Collaborative Building Task}
\label{sec:collaborative}
\paragraph{Settings.} %
In this task, the models are only trained to generate a sequence of actions without the need of considering \textit{Ask for clarifications} and \textit{Others}.
We first compare our model against the only baseline\footnote{\url{https://github.com/prashant-jayan21/minecraft-bap-models}} named BAP~\cite{jayannavar2020learning} by only using the "Execution (Original)" dataset as in Table \ref{tbl:statistics}.
Then, we conduct the experiments by using the augmented data from~\citet{jayannavar2020learning}:
the models are trained and evaluated with the augmented training data: 
5${\small,}$563 (indicated as 2x), 9${\small,}$272 (4x), and 12${\small,}$981 (6x) training samples. 
We provide the ground-truth previous actions and the world state for the next action prediction.
For the sake of fairness, we retrain the BAP model under the same setting.

For our models, we present the performance of two different dialogue context encoders: we use the pre-trained GloVe word embeddings with 300 dimensions~\cite{pennington2014glove} as the initial word embeddings followed by a GRU~\cite{chung2014empirical} 
and contextual word embeddings using the pre-trained BERT base model~\cite{devlin2018bert}. 
For the action type slot, we pre-define three potential values: \textit{placement}, \textit{removal}, and \textit{stop}. The value of the location slot can be one of 1${\small,}$089 candidate voxels and the value of the color slot can be one of six candidate colors. During training we minimize the sum of the cross entropy losses of the location slot, the color slot, and the action type slot. 
The F1-score metric on the test set is used to evaluate model performance by comparing the model predictions against the action sequence performed by the human builder. 
\paragraph{Results.} %
In Table \ref{tbl:oracle}  we present the results of our model and the baselines for the collaborative building task on the Minecraft Corpus Dataset. Experimental results show that our model outperforms the baseline model with a large margin.
Results on the augmented dataset show that the advantage of the data augmentation  
is not obvious. 
The performance using contextualized word embeddings is poorer. This could be due to the size of the builder model with the BERT encoder which makes it more difficult to train.

\begin{figure*}[!th]
  \centering
  \includegraphics[width=\textwidth]{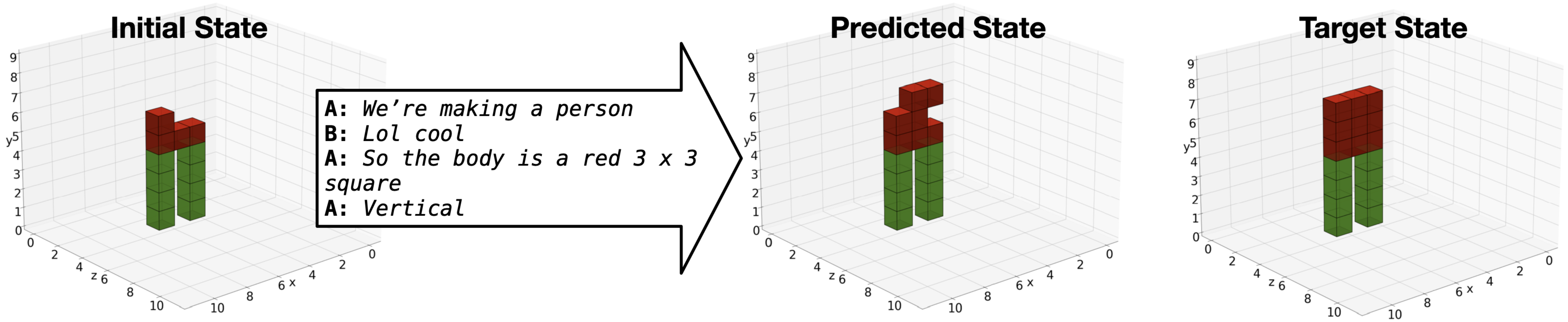}
  \caption{Case study of the collaborative building task in Sec.~\ref{sec:collaborative}: \textit{A} and \textit{B} represents the architect and the builder. 
  }
  \label{fig:case}
\end{figure*}

\subsection{Learning to Ask Task}
\label{sec:learn_to_ask}
\paragraph{Settings.} 
In this task, all the models are trained only to predict one type of actions, \textit{Execution}, \textit{Ask for clarifications} and \textit{Others}, without the need to generate a sequence of actions. All datasets in the Table \ref{tbl:statistics} are used.
In our model, for the action type slot, we define three potential slot values: \textit{Execution}, \textit{Ask}, and \textit{Others}. 
In this experiment, the slots for location and color are not used. %
We use the pre-trained GloVe embeddings in the dialogue context encoder. During the training, the cross entropy loss of the action type is minimized.
\begin{table}[ht!]
\centering
\footnotesize
\caption{Test Accuracy of the learning to ask task.}
\begin{adjustbox}{max width=\textwidth}
\begin{tabular}{lc|rrr|r}
\hline
\hline
\multicolumn{2}{c|}{\multirow{2}{*}{\begin{tabular}[c]{@{}c@{}}Test \\ Accuracy(\%)\end{tabular}}} & \multicolumn{3}{c|}{Prediction} & \multicolumn{1}{c}{\multirow{2}{*}{Size}} \\ \cline{3-5}
\multicolumn{2}{c|}{} & \multicolumn{1}{c}{Execute} & \multicolumn{1}{c}{Ask} & \multicolumn{1}{c|}{Others} & \multicolumn{1}{c}{} \\ \hline
\multicolumn{1}{l|}{\multirow{3}{*}{Oracle}} & Execution & \textbf{93.81} & 4.33 & 1.86 & 1616 \\
\multicolumn{1}{l|}{} & Ask & 22.09 & \textbf{63.80} & 14.11 & 163 \\
\multicolumn{1}{l|}{} & Others & 35.79 & 36.89 & \textbf{27.32} & 366 \\ \hline
\multicolumn{2}{c|}{Overall Test Acc} & \multicolumn{3}{c|}{80.05} & 2145 \\ \hline
\hline
\end{tabular}
\end{adjustbox}
\label{tbl:learn_to_ask}
\end{table}

\paragraph{Results.} In Table \ref{tbl:learn_to_ask}, we present the results of 
our model. 
Although our model achieves around 80\% overall test accuracy, the correct answers mainly come from the execution type while the model struggles with the ask and other types. These two types have in fact a joint test accuracy of 38.6\%. Experimental results demonstrate that the difficulty of the learning to ask task and that there is still a large room for improvement.

\subsection{Joint Learning Task}
\label{sec:joint_task}
\paragraph{Settings.} 
In this task, the models are trained to not only predict one type of actions from \textit{Execution}, \textit{Ask for clarifications} and \textit{Others} but also to generate a sequence of actions. All the datasets in Table \ref{tbl:statistics} are used.
In our model, for the action type slot, we pre-define five potential values: \textit{Placement}, \textit{Removal}, \textit{Stop}, \textit{Ask}, and \textit{Others}. The value of the location slot can be one of 1${\small,}$089 candidate grid and the value of the color slot is one of six candidate colors. We still use pre-train GloVe embedding in the dialogue context encoder. During the training we minimize the sum of the cross entropy losses of the location slot, the color slot, and the action type slot with weights equal to 0.1, 0.1, 0.8.
We provide the ground-truth previous actions and the world state for the next action prediction.
\begin{table}[ht!]
\small
\centering
\caption{Test accuracy of the joint task.}
\begin{adjustbox}{max width=\textwidth}
\begin{tabular}{cc|crr}
\hline
\hline
\multicolumn{2}{c|}{\multirow{2}{*}{\begin{tabular}[c]{@{}c@{}}Test \\ Accuracy(\%)\end{tabular}}} & \multicolumn{3}{c}{Prediction} \\ \cline{3-5} 
\multicolumn{2}{c|}{} & Execution & \multicolumn{1}{c}{Ask} & \multicolumn{1}{c}{Others} \\ \hline
\multicolumn{1}{c|}{\multirow{3}{*}{Oracle}} & Execution & \multicolumn{1}{r}{\textbf{82.64}} & 8.64 & 8.72 \\
\multicolumn{1}{c|}{} & Ask & \multicolumn{1}{r}{8.61} & \textbf{60.93} & 30.46 \\
\multicolumn{1}{c|}{} & Others & \multicolumn{1}{r}{25.47} & 47.07 & \textbf{31.46} \\ \hline
\multicolumn{2}{c|}{Overall Test Acc} & \multicolumn{3}{c}{72.26} \\ \hline
\hline
\end{tabular}
\end{adjustbox}
\label{tbl:joint_acc}
\end{table}
\begin{table}[ht!]
\small
\centering
\caption{The evaluation of the joint learning task.}
\begin{adjustbox}{max width=\textwidth}
\begin{tabular}{l|ccc}
\hline
\hline
 & F1 & Recall & Precison \\ \hline
\multicolumn{1}{c|}{Ours} & \multicolumn{1}{r}{28.4} & \multicolumn{1}{r}{20.9} & \multicolumn{1}{r}{43.9} \\ \hline
\hline
\end{tabular}
\end{adjustbox}
\label{tbl:joint_f1}
\end{table}

\paragraph{Results.} 
In Table \ref{tbl:joint_acc}, we present the results of our model's test accuracy for each action type. 
The model has an 72.3\% test accuracy. However, if the execution of building actions is excluded, its joint test accuracy of ask and other action types is about 40.5\%, indicating that deciding when to take the initiative remains challenging. 
In Table \ref{tbl:joint_f1}, we also report the results for the building task. Not surprisingly, the performance of our model drops slightly compared to those in Table \ref{tbl:oracle}, reflecting the difficulty of jointly learning.

\subsection{Case Study}

Although our model can predict the actions more accurately than the baseline, for example our model can usually predict the color of the blocks correctly with about 60\% test accuracy rate, it is still non-trivial for our model to predict the whole action sequence correctly. 
In Figure \ref{fig:case}, the architect instructed the builder to build a 3x3 square and then our model generated only parts of the structure successfully. 

The dataset noise makes the learning process more challenging: the builder action sequences are noisy due to, for example, the builder miss-clicking in the construction  process~\cite{narayan2020towards}. Also, builder action sequences are often fragmented between utterances due to the frequent interruptions of the architect. 
In order to solve these issues a good model should be capable to learn better representations for higher-level abstractions in natural language like spatial relation concepts and be more robust to noisy actions~\cite{shi2022stepgame}. 
However, existing models including pre-trained ones~\cite{devlin2018bert} fail to learn such representations for spatial reasoning, which translates into poor performance in these instruction following tasks.

\section{Conclusion}
In this paper, we extend the Minecraft Corpus dataset by labelling builder utterances into eight types, in which two of them are relevant to clarification questions. This allows builder models to learn to take initiatives in the instruction following tasks. Also, we have proposed a new model that achieves state-of-the-art performance on the Minecraft collaborative building task with a large improvement. %
Besides these contributions, we introduce a new learning to ask task for clarification questions and a jointly learning task with the collaborative building task. 
We leave the generation of clarification questions to future work.

\newpage
\bibliography{main}
\bibliographystyle{acl_natbib}

\clearpage
\appendix

\section{Appendix}
\label{sec:appendix}
\textbf{Training Details.}
We examine our model on the extend Minecraft Dialogue Corpus dataset. 
Our model's hyper-parameters are fixed for all three experiments in the Sec \ref{sec:experiment} as follows. The number of 3D-conv layers $k$ is 3, the dimension of each grid representation $d_c$ is 300, the number of layers of the grid cross-modality modules $N_G$ is 4, and the number of layers of the text cross-modality modules $N_T$ is 2. The max length of the dialogue context $s$ is selected as 100 and the dropout rates are all set to 0.2. The number of heads for the attention mechanism in the text singe and cross modality modules is set to 2, while the number of heads is set to 1 for the attention mechanism in the grid singe and cross modality modules. 
The cross entropy loss from the location slot is not counted if the ground truth label of the action type is not `placement' or 'removal', and the cross entropy loss from the color slot is not counted if the ground truth label of the action type is not 'placement'. 
The ground-truth of previous actions and the world state is  provided to the models during the training and testing.
For the experiment in the Sec \ref{sec:learn_to_ask} and \ref{sec:joint_task}, we randomly sample from 'Ask' and 'Others' sets in the training set to make training samples of different action types ('Execution', 'Ask', and 'Others') in the training set balanced. 
We train our model with cross entropy loss functions of all slots and a batch size of 50, using Adam optimizer~\cite{kingma2014adam} with a learning rate of 1e-6, $\beta_1$ = 0.9 and $\beta_2$ = 0.99. We train our model with 50 epochs and select the model with the highest F1-score on the valid set.

\end{document}